\documentclass[10pt,twocolumn,letterpaper]{article}

\usepackage[pagenumbers]{cvpr} 
\usepackage{cuted}    
\usepackage{capt-of} 

 \usepackage[most]{tcolorbox}
 \tcbset{colback=white, colframe=black!10, boxsep=2pt, left=4pt, right=4pt, top=3pt, bottom=3pt}

 \usepackage{tikz}
 \usetikzlibrary{arrows.meta}
 \usepackage{booktabs}
 \usepackage{xcolor}

 \usepackage{booktabs}
 \usepackage[table]{xcolor}
 \usepackage{siunitx}
\definecolor{bestbg}{RGB}{220,245,230}      
\definecolor{secondbg}{RGB}{235,245,255}    
\definecolor{poorbg}{RGB}{255,238,238}      

\definecolor{cvprblue}{rgb}{0.21,0.49,0.74}
\usepackage[pagebackref,breaklinks,colorlinks,allcolors=cvprblue]{hyperref}

\def\paperID{*****} 
\def\confName{CVPR}
\def\confYear{2026}

\title{QuantumCanvas: A Multimodal Benchmark for  Visual\\ Learning of Atomic Interactions}

\author{
\begin{tabular}{c c}
\begin{minipage}[t]{0.45\linewidth}\centering
{\small Can Polat}\\
{\small Texas A\&M University}\\
{\footnotesize \texttt{can.polat@tamu.edu}}
\end{minipage}
&
\begin{minipage}[t]{0.45\linewidth}\centering
{\small Erchin Serpedin}\\
{\small Texas A\&M University}\\
{\footnotesize \texttt{eserpedin@tamu.edu}}
\end{minipage}
\\[12mm]
\begin{minipage}[t]{0.45\linewidth}\centering
{\small Mustafa Kurban}\\
{\small Ankara University}\\
{\small Texas A\&M University at Qatar}\\
{\footnotesize \texttt{kurbanm@ankara.edu.tr}}
\end{minipage}
&
\begin{minipage}[t]{0.45\linewidth}\centering
{\small Hasan Kurban}\\
{\small Hamad Bin Khalifa University}\\
{\footnotesize \texttt{hkurban@hbku.edu.qa}}
\end{minipage}
\end{tabular}
}

\begin{document}
\maketitle

\begin{strip}
  \centering
  \vspace{-0.8em}
  \includegraphics[width=\linewidth]{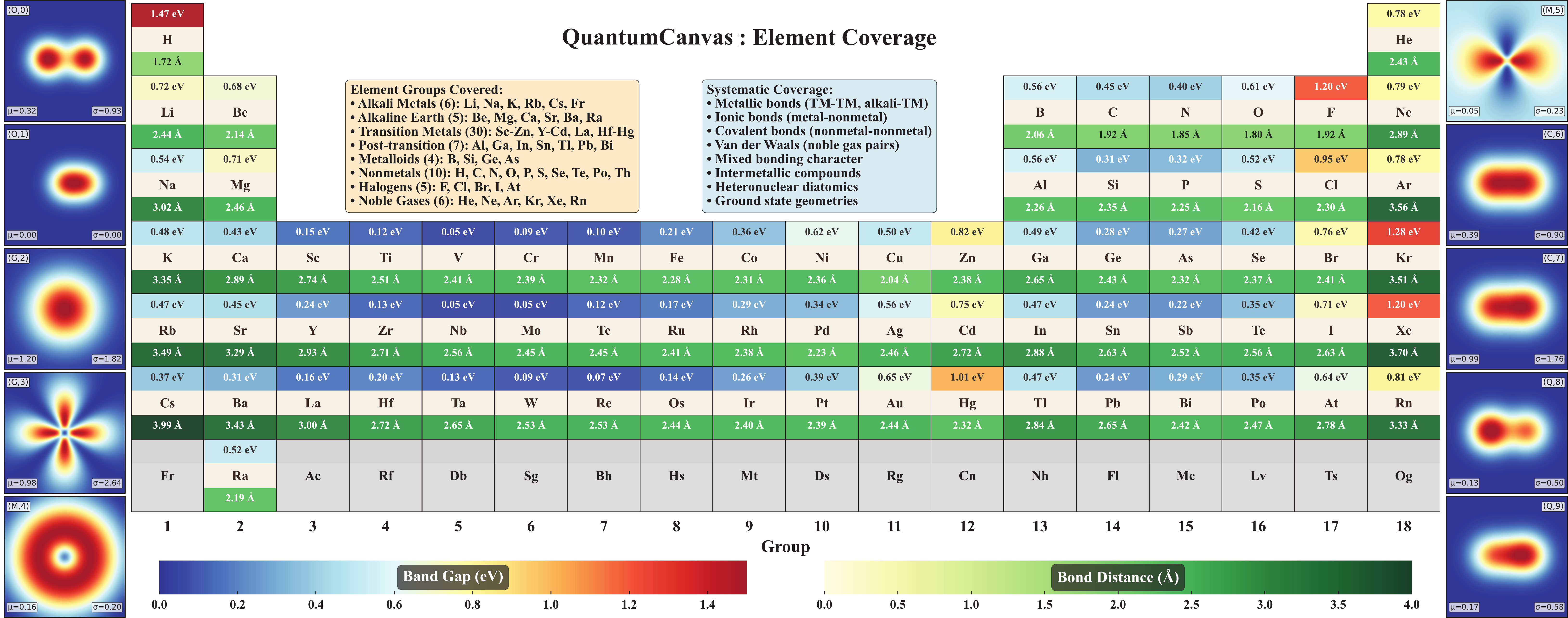}
  \captionof{figure}{Periodic-table summary for the 75 elements included in 2,850 diatomic calculations. Each cell shows the element symbol, the mean Kohn–Sham energy gap, and the mean equilibrium bond distance. The energy-gap distribution is skewed toward metallic systems (median \(0.00\) eV, IQR \(0.00\)–\(0.36\) eV), while bond distances cluster near \(2\)–\(3\)~\AA{} (median \(2.48\)~\AA{}, IQR \(2.16\)–\(2.91\)~\AA{}). The left column displays channels \(0\)–\(4\) and the right column channels \(5\)–\(9\); badges report per-channel mean \((\mu)\) and standard deviation \((\sigma)\) over all \(2{,}850\) systems.}
  \label{fig:introFig}
  \vspace{0.3em}
\end{strip}

\begin{abstract}
Despite rapid advances in molecular and materials machine learning, most models lack physical transferability: they fit correlations across whole molecules or crystals rather than learning the quantum interactions between atomic pairs. Yet bonding, charge redistribution, orbital hybridization, and electronic coupling all emerge from these two-body interactions that define local quantum fields in many-body systems. We introduce \textit{QuantumCanvas}, a large-scale multimodal benchmark that treats two-body quantum systems as foundational units of matter. The dataset spans 2,850 element–element pairs, each annotated with 18 electronic, thermodynamic, and geometric properties and paired with ten-channel image representations derived from $l$- and $m$-resolved orbital densities, angular field transforms, co-occupancy maps, and charge-density projections. These physically grounded images encode spatial, angular, and electrostatic symmetries without explicit coordinates, providing an interpretable visual modality for quantum learning. Benchmarking eight architectures across 18 targets, we report MAEs of \textbf{0.201 eV} on energy gap with GATv2, \textbf{0.265 eV} on HOMO, and \textbf{0.274 eV} on LUMO with EGNN; for energy-related quantities DimeNet attains \textbf{2.27 eV} total-energy MAE and \textbf{0.132 eV} repulsive-energy MAE, while a multimodal fusion model achieves a \textbf{2.15 eV} Mermin free-energy MAE. Pretraining on \textit{QuantumCanvas} further improves convergence stability and generalization when fine-tuned on larger dataset such as \textit{QM9}, \textit{MD17}, and \textit{CrysMTM}. By unifying orbital physics with vision-based representation learning, \textit{QuantumCanvas} provides a principled and interpretable basis for learning transferable quantum interactions through coupled visual and numerical modalities. Dataset and model implementations can be found at \url{https://github.com/KurbanIntelligenceLab/QuantumCanvas}.
\end{abstract}

\section{Introduction}
\label{sec:intro}

Materials science underpins modern advances in semiconductors, energy conversion, catalysis, and quantum technologies \cite{callister2020materials}. A central challenge lies in understanding how atomic arrangements translate into measurable electronic, thermodynamic, and structural properties \cite{giustino2014materials}. While data-driven models trained on large molecular and crystalline repositories have accelerated discovery, most remain statistical surrogates rather than physically grounded learners of interaction \cite{butler2022machine}. State-of-the-art architectures such as SchNet \cite{schutt2018schnet}, DimeNet++ \cite{gasteiger2020fast}, PaiNN \cite{schutt2021equivariant}, GemNet \cite{gasteiger2021gemnet}, Equiformer \cite{liao2022equiformer}, NequIP \cite{batzner20223}, Pure2DopeNet \cite{polat2024multimodal}, Allegro \cite{nomura2025allegro}, and GotenNet \cite{aykent2025gotennet} achieve impressive accuracy through geometric message passing and equivariant tensor updates. However, they encode atomic interactions implicitly, capturing correlations across full structures rather than the fundamental quantum mechanisms through which atoms exchange charge, form bonds, and couple orbitals \cite{unke2021machine}. As a result, their generalization often deteriorates for unseen elements, charge states, or coordination motifs, revealing a limited understanding of transferable physical principles \cite{rupp2012fast}. At the most fundamental level, bonding, charge redistribution, energy exchange, and orbital hybridization emerge from \emph{two-body quantum interactions} \cite{parr1989density}. These interactions define the local potential landscape on which all many-body effects are built. Explicitly learning them offers a direct route to physics-aware representations that are not tied to any specific geometry or composition \cite{behler2016perspective}. Despite their foundational importance, large-scale datasets systematically targeting two-body quantum systems remain scarce, leaving this essential regime largely unexplored in machine learning.

To address this gap, we introduce \textit{QuantumCanvas}, a large-scale multimodal benchmark that redefines two-body quantum systems as the elemental building blocks of matter. The dataset enumerates all 2,850 element–element combinations across 75 elements and provides eighteen complementary descriptors covering energetic, electronic, thermodynamic, and geometric properties. Each pair is also represented by a ten-channel image tensor derived from \(l\)- and \(m\)-resolved orbital densities~\cite{varshalovich1988quantum}, angular field transforms such as GAF and MTF~\cite{wu2024extracting,wang2022new}, co-occupancy maps~\cite{taulbjerg1975inner}, and charge-density projections~\cite{mulliken1933electronic,bader1981quantum,mayhall2010oniom}. These physically grounded images encode spatial, angular, and electrostatic symmetries without explicit coordinates, creating a visually interpretable modality that complements graph-based representations. All diatomics are computed under a consistent Kohn–Sham framework~\cite{bickelhaupt2000kohn,dreizler2012density} at fixed electronic temperature using Mermin’s finite-temperature formalism~\cite{stoitsov1988density}, with identical grids, smearing, and dispersion settings, ensuring that energies, orbital levels, and charge distributions remain statistically comparable across pairs. The benchmark evaluates eight representative architectures across eighteen targets using mean absolute error under both random and composition-held-out splits, averaged over three seeds with uniform hyperparameter budgets. Pretraining on \textit{QuantumCanvas} yields transferable embeddings that improve accuracy, convergence, and robustness when fine-tuned on established molecular and crystalline benchmarks such as \textit{QM9} \cite{ramakrishnan2014quantum}, \textit{MD17} \cite{chmiela2017machine}, and \textit{CrysMTM} \cite{polat2025crysmtm}. These results demonstrate that representations learned from two-body quantum interactions provide a physically meaningful foundation for generalizing across molecular, dynamical, and solid-state regimes.

The main contributions of this work are: (1) a comprehensive dataset of 2,850 diatomic systems covering 75 elements with 18 quantum-mechanical descriptors; (2) a ten-channel coordinate-free image representation derived from orbital populations and charge features that encodes symmetry for vision and multimodal learning; (3) a unified benchmark of eight architectures across eighteen targets under matched budgets, revealing modality-specific inductive biases and a well-defined difficulty gradient; and (4) experimental evidence that two-body pretraining improves accuracy, convergence, and generalization on \textit{QM9}, \textit{MD17}, and \textit{CrysMTM}. Figure~\ref{fig:introFig} summarizes the elemental coverage and per-channel statistics that motivate this multimodal design.

\section{Related Work}
\label{sec:relatedWork}

\noindent
\textbf{Modeling atomic interactions.}  
In quantum mechanics, all molecular and material phenomena originate from interactions between atoms \cite{levine2009quantum}. The two-body problem forms the foundation of quantum chemistry, governing the potential energy and charge distribution between a pair of nuclei and their electrons \cite{atkins2011molecular}. Classical models such as the Lennard–Jones and Morse potentials approximate these interactions parametrically \cite{israelachvili2011intermolecular, varshni1957comparative}, while first-principles approaches like density functional theory (DFT) resolve them directly from the electronic structure \cite{martin2020electronic}. In condensed-matter physics, pair potentials remain indispensable for modeling cohesive energies, lattice stability, and defect formation \cite{mahan2013many}. Despite their universality, explicit representations of two-body quantum behavior are largely absent from modern machine learning frameworks \cite{ramakrishnan2017machine}. Most architectures encode atomic neighborhoods as high-dimensional embeddings with limited physical interpretability \cite{denell2025automated}. Capturing two-body interactions explicitly is critical for achieving transferability, as these interactions form a universal basis that generalizes across molecular, crystalline, and amorphous systems \cite{airas2023transferable, yamada2019predicting}.

\noindent
\textbf{Existing datasets in molecular and materials machine learning.}  
The rapid progress of materials informatics has been enabled by large-scale quantum datasets. Benchmarks such as QM7-X \cite{hoja2021qm7} and MD22 \cite{chmiela2023accurate} capture molecular geometries and DFT-level electronic properties for diverse organic systems, while the Materials Project, OQMD \cite{saal2013materials}, NOMAD \cite{draxl2019nomad}, and Matbench \cite{dunn2020benchmarking} provide millions of crystalline and surface calculations. These resources have catalyzed the development of graph-based and equivariant models that enable data-driven discovery of materials and molecules \cite{de2015database}. However, they primarily describe fully realized molecular or bulk systems, where pairwise interactions are entangled with many-body effects, polarization, and geometry-dependent coupling \cite{ibragimova2025unifying}. As a result, they do not isolate the primitive two-body quantum observables that underpin all higher-order material behavior \cite{amico2008entanglement}.

\noindent
\textbf{Two-body datasets and their limitations.}  
Only a few specialized efforts have targeted atomic pair interactions, often in the context of potential fitting and  spectroscopy \cite{patkowski2020recent}. Frameworks such as ANI \cite{smith2017ani} and MACE \cite{batatia2022mace} incorporate pairwise contributions through local energy decomposition but lack explicit two-body reference data. Experimental and \textit{ab initio} diatomic datasets (e.g., NIST diatomic constants) provide limited elemental coverage and omit uniform descriptors beyond energy–distance curves \cite{werner2020molpro}. Other benchmarks such as OC20 \cite{chanussot2021open} and OC22 \cite{tran2023open} emphasize many-body reaction energetics and surface adsorption, focusing on chemical processes rather than isolated atomic physics. Consequently, no existing benchmark systematically enumerates element–element combinations while providing both scalar quantum properties and spatially resolved orbital representations \cite{charkin2025atomic}.

\noindent
\textbf{Positioning of QuantumCanvas.}  
\textit{QuantumCanvas} addresses this gap by modeling two-body quantum systems as the fundamental representational units of matter \cite{christensen2025machine}. It unifies numerical descriptors with image-based orbital modalities that encode spatial, angular, and electrostatic symmetries \cite{zhang2023universal}. This multimodal design provides an interpretable and physically consistent foundation for learning interatomic physics through both numerical and visual modalities, bridging quantum chemistry, computer vision, and materials informatics \cite{polat2025xchemagents}. By doing so, \textit{QuantumCanvas} establishes the first large-scale benchmark that connects orbital-level quantum behavior to modern representation learning.

\section{Dataset}
\label{sec:dataset}

\subsection{Two-body Quantum Simulations}
\noindent

Each sample in \textit{QuantumCanvas} corresponds to an isolated diatomic (A--B) system computed through a consistent quantum-chemistry workflow based on an approximate finite-temperature Kohn--Sham density functional framework. Within the Mermin formalism, the electronic free energy is written as
\begin{equation}
\label{eq:mermin_free_energy}
\mathcal{F}[\rho] \;=\; E_\mathrm{KS}[\rho] \;-\; T_e\,S[\{f_i\}],
\end{equation}
where
\begin{equation}
\label{eq:E_KS}
E_\mathrm{KS}[\rho] \;=\; T_s[\rho] \;+\; E_\mathrm{H}[\rho] \;+\; E_\mathrm{xc}[\rho] \;+\; \int \rho(\mathbf{r})\,v_\mathrm{ext}(\mathbf{r})\,d\mathbf{r},
\end{equation}
and \(S\) is the Fermi–Dirac entropy of the occupations \(\{f_i\}\). Reported energy components include the band (one-electron) energy
\begin{equation}
\label{eq:E_band}
E_\mathrm{band} \;=\; \sum_i f_i\,\varepsilon_i,
\end{equation}
the short-range repulsion \(E_\mathrm{rep}\), and the Mermin free energy evaluated at the self-consistent density \(\rho^\star\) \cite{ziman1979principles}. We record
\begin{equation}
\label{eq:E_tot_F}
E_\mathrm{tot} \;=\; E_\mathrm{band} \;+\; E_\mathrm{rep} \;+\; \Delta E,
\qquad
F \;=\; E_\mathrm{tot} \;-\; T_e S,
\end{equation}
where \(\Delta E\) collects exchange–correlation and double-counting corrections \cite{kresse1996efficient, aradi2007dftb+}. Geometry optimization proceeds via SCF cycles to tolerance \(\varepsilon_\mathrm{SCF}\) and line-search relaxation of the separation \(r=\|\mathbf{R}_B-\mathbf{R}_A\|\) until \(\bigl|\,\partial E_\mathrm{tot}/\partial r\,\bigr|<\varepsilon_\mathrm{geom}\) \cite{nocedal2006numerical}. We also record the Fermi level \(E_F\), the dipole moment
\begin{equation}
\label{eq:dipole}
\boldsymbol{\mu} \;=\; \sum_{\alpha\in\{A,B\}} q_\alpha\,\mathbf{R}_\alpha \;-\; \int \mathbf{r}\,\rho(\mathbf{r})\,d\mathbf{r},
\end{equation}
and Mulliken-like gross atomic charges \(q_\alpha\), together with orbital populations resolved by angular momentum \((\ell,m)\) from projected density matrices. 

All quantum-chemical calculations are carried out within the self-consistent charge density-functional tight-binding (SCC-DFTB) framework \cite{gaus2011dftb3}, which can be viewed as an approximate realization of finite-temperature Kohn--Sham density functional theory, as implemented in the DFTB+ code~\cite{aradi2007dftb+}. The Slater--Koster matrix elements were taken from the periodic table baseline parameters (PTBP) set, which provides robust density functional tight-binding parameters for solids across the periodic table and covers all elements considered in this work~\cite{cui2024ptbp}. This level of theory offers a good compromise between accuracy and computational cost, enabling the large-scale simulations.

\subsection{Image Representations from Orbital Populations}
\noindent
Beyond scalars, each diatomic system is encoded as a fixed-size 10-channel image tensor \(\mathbf{I}\in\mathbb{R}^{10\times 32\times 32}\) derived from \((\ell,m)\)-resolved orbital populations. Let \(n_{\ell m}^{(\alpha)}\) denote the population on atom \(\alpha\in\{A,B\}\) in the basis of real spherical harmonics \(Y_{\ell m}\). We construct the following modalities.

\paragraph{Orientation-aware orbital map (O-Map).}
\noindent
For each atom, we rasterize orbital populations over indices \((\ell,m)\) up to \(\ell_\mathrm{max}=2\) (covering \(s\), \(p\), and \(d\) shells):
\begin{equation}
\label{eq:omap}
\mathrm{OMap}^{(\alpha)}[\ell,m] \;=\; n_{\ell m}^{(\alpha)}.
\end{equation}
Each map is padded to a \(3\times 5\) tile on a \(32\times 32\) canvas; the two atomic maps form channels \(\mathbf{I}_{0:2}\).

\paragraph{Rotationally invariant power over shells (RIP) with GAF and MTF.}
\noindent
Shell populations \(P_\ell^{(\alpha)}=\sum_{m=-\ell}^{\ell} n_{\ell m}^{(\alpha)}\) are normalized to \([0,1]\) to obtain \(\tilde{P}_\ell^{(\alpha)}\). The Gramian Angular Field (GAF) maps \(\tilde{\mathbf{P}}^{(\alpha)}=[\tilde{P}_0,\tilde{P}_1,\tilde{P}_2]\) to
\begin{equation}
\label{eq:gaf}
\mathrm{GAF}^{(\alpha)}[i,j] \;=\; \cos\!\bigl(\phi_i+\phi_j\bigr),
\end{equation}
with angular encoding
\begin{equation}
\label{eq:phi}
\phi_i \;=\; \arccos \tilde{P}^{(\alpha)}_i.
\end{equation}
The Markov Transition Field (MTF) quantizes \(\tilde{\mathbf{P}}^{(\alpha)}\) into \(Q=3\) states by rank and encodes co-occurrence via
\begin{equation}
\label{eq:mtf}
\mathrm{MTF}^{(\alpha)}[u,v] \;=\; \mathbb{P}\!\bigl(s_{t+1}=v \,\big|\, s_t=u\bigr),
\end{equation}
for \(u,v\in\{1,2,3\}\). Both fields are upsampled to \(32\times 32\) and assigned to channels \(\mathbf{I}_{2:4}\) (GAF) and \(\mathbf{I}_{4:6}\) (MTF).

\paragraph{Co-occupancy map (COM).}
\noindent
Inter-atomic shell coupling is captured by the outer product of shell sums:
\begin{equation}
\label{eq:com}
\mathrm{COM}[\ell_1,\ell_2] \;=\; P_{\ell_1}^{(A)}\,P_{\ell_2}^{(B)}, \qquad \ell_1,\ell_2\in\{0,1,2\}.
\end{equation}
This map is upsampled to \(32\times 32\) and stored as channel \(\mathbf{I}_{6}\). An optional second COM channel \(\mathbf{I}_{7}\) encodes normalized shell distributions or population-weighted \(|m|\) moments.

\paragraph{Charge images (Q-Image).}
\noindent
Charge features are summarized by three \(2\times 2\) matrices:
\(\mathbf{Q}_{\mathrm{diag}}=\begin{bmatrix}q_A&0\\0&q_B\end{bmatrix},
\mathbf{Q}_{|\Delta|}=\begin{bmatrix}0&|q_A-q_B|\\|q_A-q_B|&0\end{bmatrix},
\mathbf{Q}_{\Pi}=\begin{bmatrix}0&q_Aq_B\\q_Aq_B&0\end{bmatrix}\),
which are bilinearly upsampled to \(32\times 32\) and mapped to channels \(\mathbf{I}_{8:10}\). The final tensor \(\mathbf{I}\) integrates orientation-aware, rotationally invariant, and electrostatic information for multimodal learning.

\subsection{Targets and Labels}
\noindent
Each pair is annotated with electronic, energetic, geometric, and charge-derived descriptors. Electronic observables include \(E_\mathrm{band}\), \(E_\mathrm{rep}\), \(E_\mathrm{tot}\), and the Mermin free energy \(F\). Frontier levels \(\varepsilon_\mathrm{HOMO}\) and \(\varepsilon_\mathrm{LUMO}\) define the gap \(E_g=\varepsilon_\mathrm{LUMO}-\varepsilon_\mathrm{HOMO}\) and the Fermi level \(E_F\). Thermodynamic quantities comprise the ionization potential \(I\) and electron affinity \(A\), from which the following conceptual-DFT descriptors are derived:
\begin{equation}
\label{eq:chem_descriptors}
\begin{gathered}
\chi = \frac{I + A}{2}, \qquad
\eta = \frac{I - A}{2}, \qquad
S = \eta^{-1}, \\[4pt]
\mu_{chem} = -\chi, \qquad
\omega = \frac{\mu^{2}}{2\eta}.
\end{gathered}
\end{equation}
where \(\chi\) denotes electronegativity, \(\eta\) hardness, \(S\) softness, \(\mu\) chemical potential, and \(\omega\) electrophilicity (all in eV) \cite{szabo2012modern}. Dipole properties include the Cartesian components \(\boldsymbol{\mu}=(\mu_x,\mu_y,\mu_z)\) and the magnitude \(\|\boldsymbol{\mu}\|\). The equilibrium bond length \(r\) (in \AA) is extracted from optimized geometries. Charge statistics \cite{richard1990atoms} summarize redistribution as follows:
$q_\mathrm{maxabs} = \max(|q_A|, |q_B|)$, 
$q_\mathrm{absmean} = \tfrac{1}{2}(|q_A| + |q_B|)$, and 
$q_\mathrm{std} = \sqrt{\tfrac{1}{2}\sum_{\alpha\in\{A,B\}} (q_\alpha - \bar q)^2}$ 
with $\bar q = \tfrac{1}{2}(q_A + q_B)$.
\noindent
\paragraph{Label computation and units.}
\noindent
Energies and orbital levels are reported in eV after conversion from Hartree (\(1\,\mathrm{Ha}=27.2114\,\mathrm{eV}\)). Dipoles are in Debye, charges in elementary charge \(e\), and distances in \AA. All values are obtained under a unified SCF and geometry protocol to ensure statistical consistency across pairs. Each record includes SCF diagnostics (iteration counts and residuals) to identify unconverged cases.

\begin{figure*}[ht]
  \centering
  \includegraphics[width=1\linewidth]{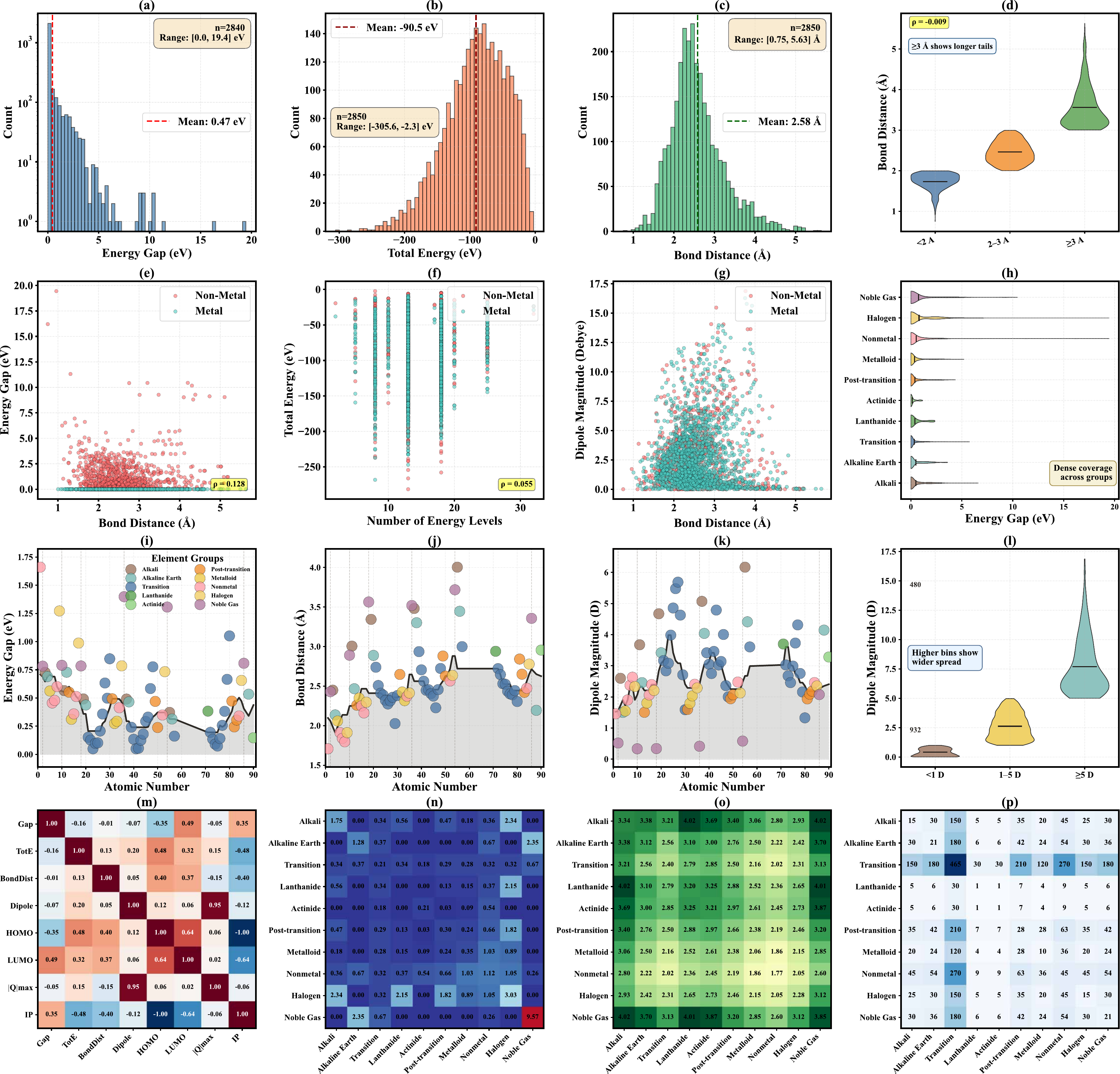}
  \caption{{Comprehensive overview of the \textit{QuantumCanvas} diatomic corpus. Panels (a)--(c) summarize the global distributions of energy gap, total energy, and bond length (log-scaled in (a)) while reporting sample coverage. Panel (d) provides a categorical breakdown of bond-length regimes, and panels (e)--(g) chart structure--property relationships for energy gap, total energy, and dipole magnitude. Panel (h) visualizes the distribution of energy gaps across elemental groups, whereas panels (i)--(k) present element-resolved “constellation’’ views of average gap, bond length, and dipole magnitude. Panel (l) stratifies dipole magnitudes into physically meaningful bins. The bottom row consolidates aggregate statistics: panel (m) displays the property correlation matrix, and panels (n)--(p) highlight the average energy gap, average bond length, and sample counts for each elemental group. Together, these subfigures illustrate the breadth, balance, and cross-property variability captured in {QuantumCanvas}, underscoring its value for benchmarking data-driven quantum chemistry models.}}
  \label{fig:statistics}
\end{figure*}

\subsection{Analysis}

\paragraph{Periodic coverage and chemical diversity.}
\noindent
Figure~\ref{fig:introFig} visualizes elemental and pairwise coverage for \textit{QuantumCanvas}, summarizing \(2{,}850\) diatomics across \(75\) elements. For each element, the periodic table reports the mean Kohn–Sham band gap and equilibrium bond distance, yielding a compact map of physical diversity. Coverage includes nearly the entire s-, p-, and d-blocks: alkali (5/6), alkaline earth (6/6), transition (30/30), post-transition (7/7), metalloid (4/4), nonmetal (10/10), halogen (5/5), and noble gas (6/6). The overall band-gap distribution (\(\mu=0.47\) eV, \(\sigma=1.19\) eV) is sharply skewed toward metallic and semi-metallic behavior, while bond distances (\(\mu=2.58\)~\AA{}, \(\sigma=0.66\)~\AA{}) concentrate in the \(2\)–\(3\)~\AA{} range. \emph{Implication: the dataset supports evaluation across chemically distinct regimes while maintaining realistic class balance for downstream generalization.}

\noindent
Channel-wise statistics in Figure~\ref{fig:introFig} show complementary signal profiles. O-Maps (\textbf{ch0–ch1}) have low means/variances, reflecting sparsity in \((\ell,m)\)-resolved populations and localized occupations. Gramian Angular Fields and Markov Transition Fields (\textbf{ch2–ch5}) exhibit the highest mean intensities and dispersions, capturing angular correlations and transition structure among \(s\)-, \(p\)-, and \(d\)-shell populations. Co-occupancy maps (\textbf{ch6–ch7}) display moderate-to-high dispersion driven by cross-atomic shell coupling that varies with the chemical pair. Charge-derived channels (\textbf{ch8–ch9}) show low-to-moderate means with broad variance, encoding charge asymmetry and multiplicative charge interactions.

\paragraph{Pairwise trends and structure–property patterns.}
\noindent
Pair counts range from \(10\) metalloid–metalloid examples to \(465\) transition–transition dimers, reflecting natural chemical prevalence and emphasis on metallic bonding. Average band gaps vary from \(0.00\) eV in alkali–alkaline-earth or noble-gas/post-transition pairs to \(\sim 9.6\) eV in noble-gas homodimers, spanning localized to delocalized regimes. Mean bond distances range from \(1.9\)~\AA{} in nonmetal–nonmetal pairs to \(4.0\)~\AA{} for weakly bound alkali–noble-gas pairs. \emph{Implication: the dataset covers covalent, metallic, and van der Waals regimes, enabling stress-tests of extrapolation across coordination and bonding types.}

\noindent
Figure~\ref{fig:statistics} provides a comprehensive characterization of the \textit{QuantumCanvas} diatomic corpus. Panels~(a)--(c) summarize the global distributions of three core scalar properties. The energy-gap distribution in~(a) is heavily right-skewed (mean $0.47$~eV, median $0$~eV, max $19.45$~eV), reflecting the broad range of excitation behaviors across the periodic table. Total energies in~(b) span over $300$~eV, from weakly bound species to strongly attractive dimers, while bond distances in~(c) cluster around $2.58$~\AA\ but still extend from extremely short ($0.75$~\AA) to long-range ($5.6$~\AA) separations. Panel~(d) further partitions bond lengths into short ($<2$~\AA, $16\%$), medium ($2$--$3$~\AA, $62\%$), and long-range ($>3$~\AA, $22\%$) regimes, showing that the dataset emphasizes chemically meaningful equilibrium-length dimers but preserves substantial coverage of compressed and stretched configurations.

Panels~(e)--(g) explore structure–property relationships. The energy gap shows essentially no dependence on bond distance~(e; $\rho=-0.009$), indicating that orbital-level excitations are governed primarily by chemical identity rather than geometry alone. Total energy increases mildly with system size proxy~(f; $\rho=0.128$), consistent with scaling of electron count, and dipole magnitude exhibits only a weak geometric trend~(g; $\rho=0.055$), confirming that polarity is mostly driven by electronegativity contrast rather than internuclear separation.

Panel~(h) organizes energy-gap distributions by periodic-table groups. Transition-metal dimers—the largest subset—show relatively small average gaps ($0.32$~eV), whereas halogen and noble-gas combinations yield broad, high-gap distributions (up to $19$~eV), revealing strong periodic trends. Panels~(i)--(k) provide element-level “constellation’’ summaries of average gap, bond length, and dipole magnitude across 75 elements. Hydrogen forms the shortest and smallest-gap dimers, heavy alkali metals form the longest bonds, and dipole magnitudes peak for highly asymmetric pairs involving Cs, illustrating chemically interpretable patterns that a model must generalize across.

Panel~(l) stratifies dipoles into physically meaningful categories: nearly one-third of dimers are weakly polar ($<1$~D), half lie in the moderate-polarity regime (1–5~D), and a significant tail ($17\%$) exceeds 5~D. This variation is crucial for evaluating whether models capture subtle charge-transfer behavior across the periodic table.

The correlation matrix in panel~(m) shows that most QuantumCanvas properties are only weakly correlated, confirming that the benchmark spans diverse and largely independent physical regimes. HOMO, LUMO, and gap form a coherent electronic cluster (e.g., $\rho_{\mathrm{HOMO,LUMO}}=0.642$), while strong dipole–charge coupling ($\rho=0.950$ between dipole and $|Q|_{\max}$) reflects consistent charge-asymmetry physics. Conversely, geometry (bond distance) correlates only moderately with frontier energies ($\rho\sim0.36$–$0.40$) and very weakly with the gap, underscoring that the dataset does not collapse into a single geometric or energetic trend. \emph{Implication: benchmarks should report per-target robustness and prioritize architectures that can disentangle electronic from geometric cues (e.g., equivariant graphs plus electronic-context heads).}

Finally, panels~(n)--(p) present group-pair statistics. Average energy gaps vary dramatically across element-group combinations—from nearly metallic (alkali–alkaline-earth pairs, $0$~eV) to highly insulating (noble-gas–noble-gas pairs, $9.57$~eV). Group-level bond lengths in~(o) range from compact (nonmetal–nonmetal, $1.77$~\AA) to extremely weakly bound (alkali–noble-gas, $4.02$~\AA). Pair-count distributions in~(p) highlight chemical diversity: transition–transition pairs dominate with up to 465 samples, while rare combinations (e.g., lanthanide–lanthanide) remain minimally represented but non-zero.  

Together, subfigures~(a)--(p) reveal that \textit{QuantumCanvas} is broad, chemically diverse, and structurally rich, with low redundancy across targets and strong periodic trends—providing an extensive and challenging benchmark for learning robust, transferable two-body quantum interactions.

\begin{figure}[h]
    \centering
    \includegraphics[width=1\linewidth]{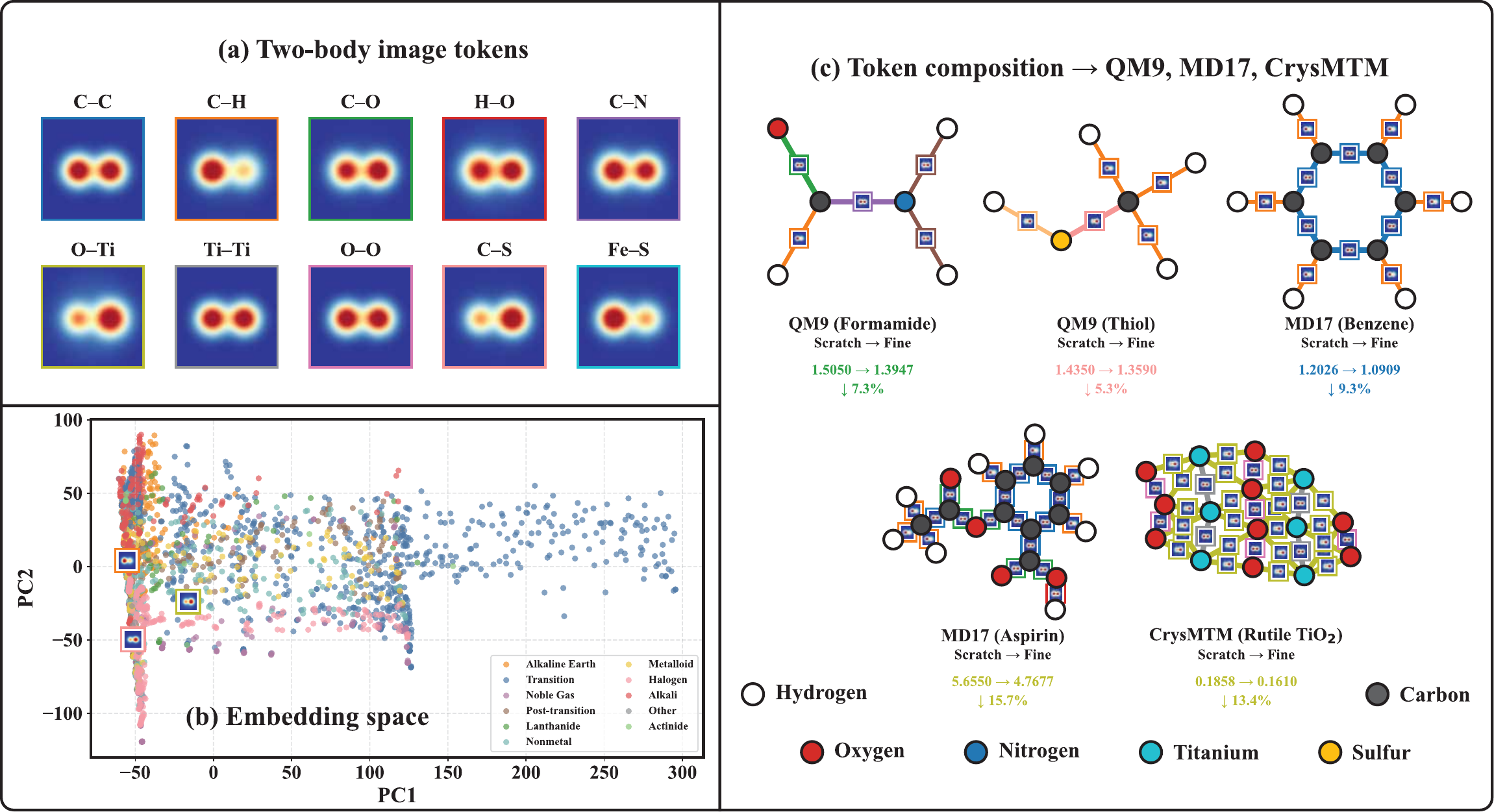}
    \caption{Panels (a–c) summarize the construction, organization, and downstream use of the two-body interaction tokens. (a) Representative two-body image tokens for selected element pairs; each inset is bordered by the color assigned to that pair type. (b) Two-dimensional PCA projection of all token embeddings, colored by the periodic-table group of one constituent element, showing that chemically related pairs cluster in the embedding space. (c) Assembly of two-body tokens into full molecular and crystalline structures for selected examples from QM9, MD17, and CrysMTM. For each bond, the corresponding token thumbnail is placed at the bond midpoint, illustrating how pairwise interactions compose into larger systems. Atoms are drawn with simplified CPK-like colors: carbon (dark gray), hydrogen (white), oxygen (red), nitrogen (blue), sulfur (yellow), and titanium (teal). Reported mean absolute errors show the improvement obtained when models are initialized from the two-body representation rather than trained from scratch.}
    \label{fig:transferLearning}
\end{figure}
\section{Experiments}
\label{sec:experiments}

\paragraph{Setup.}
\noindent
We evaluate learning of two-body quantum interactions on \textit{QuantumCanvas}, which spans 18 scalar targets across electronic, energetic, thermodynamic, and geometric categories (Table~\ref{tab:qc_full}). Each sample provides two modalities: a graph branch with atomic numbers and 3D coordinates, and a vision branch with ten-channel images derived from \((\ell,m)\)-resolved orbital populations, angular transforms, and charge maps. Models are trained with mean absolute error (MAE), early stopping on a held-out validation split, and three random seeds; we report mean \(\pm\) standard deviation. Splits are \emph{element–pair disjoint} to prevent information leakage (i.e., a given A–B pair never appears across train/val/test), and preserve the class balance reported in Figure~\ref{fig:statistics}. Unless stated otherwise, methods receive identical inputs and differ only by encoder and fusion strategy. Implementation details are provided in the supplementary material.

\paragraph{Baselines and modalities.}
\noindent
We compare geometric/equivariant graph networks (SchNet, DimeNet \cite{gasteiger2020directional}, EGNN \cite{satorras2021n}, FAENet \cite{duval2023faenet}, GATv2 \cite{brody2021attentive}) to vision encoders trained on images alone (ViT \cite{dosovitskiy2020image}, QuantumShellNet \cite{polat2025quantumshellnet}) and a late-fusion model that concatenates graph and image features before prediction. Hyperparameters (depth, width, cutoffs, radial/angle bases) are tuned uniformly per family to avoid target-specific bias.

\paragraph{Difficulty landscape across targets.}
\noindent
Targets exhibit a clear complexity gradient. Charge statistics \((Q_{\max}, Q_{\text{mean}}, Q_{\text{std}})\) and equilibrium bond distance achieve the lowest MAE, reflecting dependence on atomic identity and pair geometry. Band and total energies show larger errors, consistent with many-electron and delocalized behavior that is only partially resolvable from two-body data. Frontier levels (HOMO, LUMO) and the gap are intermediate. Conceptual DFT quantities (electronegativity, chemical potential, hardness, softness, electrophilicity) are most sensitive to numerical stability due to non-linear transforms.

\paragraph{Modality analysis: graph vs. image vs. fusion.}
\noindent
Graph encoders excel on geometry- and distance-driven properties (e.g., \textsc{$r$}, \textsc{$\|\boldsymbol{\mu}\|$}). Image encoders capture complementary electronic patterns from orbital symmetry, shell populations, and charge anisotropy. QuantumShellNet (image-only) approaches graph performance on several electronic targets, showing that orbital-field images contain rich quantum information even without explicit geometry. Late fusion improves selected properties such as HOMO, LUMO, and $\|\boldsymbol{\mu}\|$, indicating synergistic rather than redundant cues. \emph{Effect sizes:} for the energy gap $E_g$, GATv2 ($0.201\pm0.020$) improves MAE by $\mathbf{11\%}$ over EGNN ($0.226\pm0.015$) and $\mathbf{19\%}$ over DimeNet ($0.248\pm0.020$); for $\|\boldsymbol{\mu}\|$, EGNN ($0.588\pm0.039$) reduces MAE by $\mathbf{7.8\%}$ vs.\ DimeNet ($0.638\pm0.035$).

\paragraph{Architectural trends.}
\noindent
Among graph models, DimeNet achieves the lowest errors on geometry- and energy-sensitive tasks, benefiting from explicit radial/angle bases (e.g., \textsc{$r$} $0.008\pm0.001$~\AA). EGNN is competitive with lower capacity, highlighting the efficiency of equivariant design (best on \textsc{$\|\boldsymbol{\mu}\|$} $0.588\pm0.039$ and tied-best/second on several electronic targets). FAENet favors directional observables (e.g., \textsc{$\mu_z$} $0.888\pm0.091$). On the vision side, QuantumShellNet outperforms ViT on orbital/image-driven targets, supporting the value of physics-aware encodings. \emph{Effect sizes:} HOMO MAE improves by $\mathbf{11.9\%}$ from DimeNet ($0.301\pm0.013$) to EGNN ($0.265\pm0.014$); LUMO improves by $\mathbf{17.2\%}$ from DimeNet ($0.331\pm0.027$) to EGNN ($0.274\pm0.032$). DimeNet reduces \textsc{Total Energy} MAE by $\mathbf{28.8\%}$ vs.\ EGNN (2.27 vs.\ 3.19~eV) and $\mathbf{11.7\%}$ vs.\ FAENet (2.27 vs.\ 2.57~eV). For \textsc{Repulsive Energy}, DimeNet (0.132~eV) improves $\mathbf{41.6\%}$ over the second-best GATv2 (0.226~eV).

\paragraph{Variance, stability, and robustness across properties.}
\noindent
Different quantum targets exhibit markedly different levels of seed sensitivity. Among the conceptual DFT descriptors, \emph{Softness} ($S$) and \emph{Electrophilicity} ($\omega$) show the largest variance across runs, reflecting their nonlinear dependence on ionization potential and electron affinity; small perturbations in these upstream quantities propagate multiplicatively, amplifying fluctuations. Mermin free energy also displays elevated variability due to its explicit entropic term, which is harder to regress from static two-body fields. In contrast, geometric observables such as the equilibrium bond length $r$ and charge statistics ($q_\mathrm{maxabs}$, $q_\mathrm{absmean}$, $q_\mathrm{std}$) are substantially more stable, with standard deviations typically below $10^{-2}$.

Electronic-level predictions exhibit moderate but consistent variance patterns across GNNs and ViT-like models, while DimeNet and EGNN remain comparatively stable owing to their inductive biases. These metrics highlight which properties are robustly learnable from two-body interactions and which are intrinsically noisy due to higher-order coupling or thermodynamic contributions.

\subsection{Transfer Learning Across Molecular and Crystalline Domains}
\noindent
We pretrain on \textit{QuantumCanvas} and fine-tune on QM9, MD17, and CrysMTM to test transfer across complementary regimes. QM9 probes molecular orbital energies, MD17 evaluates force-consistent potential energy surfaces along dynamics trajectories, and CrysMTM extends to periodic solids with composition and temperature effects. Pretraining on two-body interactions accelerates convergence and improves validation stability relative to training from scratch. Figure~\ref{fig:transferLearning} visualizes the two-body interaction tokens, their embedding structure, and their assembly into downstream molecular and crystalline systems, highlighting how these representations bootstrap learning across datasets. On QM9, initialization reduces errors for LUMO and the gap; on MD17 it improves Aspirin, Benzene, and Ethanol targets; on CrysMTM it facilitates learning composition- and temperature-dependent band-gap variations with reduced overfitting. These results indicate that two-body pretraining internalizes low-order electronic principles—orbital coupling, charge asymmetry, short-range curvature—that transfer to many-body and periodic settings, supporting two-body pretraining as a scalable foundation for molecular and materials learning.

\begin{table}[t]
\centering
\footnotesize
\setlength{\tabcolsep}{3pt}
\caption{\textbf{Comprehensive Benchmark Results on \textit{QuantumCanvas}.} 
MAE $\pm$ standard deviation across 18 target properties and eight architectures. Units follow the dataset conventions: energies and orbital levels in eV, dipoles in Debye, charges in elementary charge $e$, and distances in \AA.
Green cells with bold values indicate the best performance, while blue underlined values denote the second-best. 
Graph-based models use atomic coordinates and chemical features, 
image-based models operate on 10-channel orbital-derived images, 
and the MultiModal model fuses both representations. Lower MAE indicates better predictive accuracy.}
\label{tab:qc_full}
\resizebox{\linewidth}{!}{%
\begin{tabular}{lcccccccc}
\toprule
Target & DimeNet & EGNN & FAENet & GATv2 & MultiModal & QShellNet & SchNet & ViT \\
\midrule
\multicolumn{9}{c}{\emph{Electronic Properties (eV)}} \\
\midrule
$E_g$ & \underline{\cellcolor{secondbg}}0.248±0.020 & \underline{\cellcolor{secondbg}}0.226±0.015 & 0.420±0.070 & \cellcolor{bestbg}\textbf{0.201±0.020} & 0.639±0.041 & \cellcolor{poorbg}1.06±0.30 & 0.662±0.11 & 0.474±0.05\\
$E_{HOMO}$ & \underline{\cellcolor{secondbg}}0.301±0.013 & \cellcolor{bestbg}\textbf{0.265±0.014} & 0.363±0.013 & 0.306±0.026 & 0.398±0.019 & \cellcolor{poorbg}0.883±0.20 & 0.496±0.02 & 0.868±0.03\\
$E_{LUMO}$ & \underline{\cellcolor{secondbg}}0.331±0.027 & \cellcolor{bestbg}\textbf{0.274±0.032} & 0.507±0.028 & \underline{\cellcolor{secondbg}}0.322±0.050 & 0.551±0.059 & 0.894±0.06 & 0.666±0.08 & 0.882±0.02\\
Energy gap & \cellcolor{bestbg}\textbf{3.34±0.10} & 3.74±0.19 & 3.83±0.40 & \cellcolor{poorbg}8.56±1.45 & \underline{\cellcolor{secondbg}}3.34±0.42 & \cellcolor{poorbg}19.4±3.1 & 4.34±0.46 & \cellcolor{poorbg}17.5±1.3\\
\midrule
\multicolumn{9}{c}{\emph{Energy Properties (eV)}}\\
\midrule
$E_{tot}$ & \cellcolor{bestbg}\textbf{2.27±0.59} & 3.19±0.52 & \underline{\cellcolor{secondbg}}2.57±0.17 & \cellcolor{poorbg}9.23±1.50 & 3.00±0.77 & \cellcolor{poorbg}19.5±1.29 & 3.03±0.47 & \cellcolor{poorbg}16.5±0.39\\
$E_{rep}$ & \cellcolor{bestbg}\textbf{0.132±0.020} & 0.244±0.019 & 0.245±0.017 & \underline{\cellcolor{secondbg}}0.226±0.001 & 0.247±0.081 & \cellcolor{poorbg}0.769±0.19 & 0.293±0.02 & 0.397±0.04\\
Mermin $E$ & 3.22±2.05 & \underline{\cellcolor{secondbg}}2.68±0.24 & 2.73±0.53 & \cellcolor{poorbg}8.17±0.88 & \cellcolor{bestbg}\textbf{2.15±0.38} & \cellcolor{poorbg}18.9±3.8 & 3.36±0.17 & \cellcolor{poorbg}16.7±0.48\\
\midrule
\multicolumn{9}{c}{\emph{Conceptual DFT Descriptors}}\\
\midrule
IP & \cellcolor{bestbg}\textbf{0.276±0.008} & \underline{\cellcolor{secondbg}}0.276±0.007 & 0.335±0.024 & 0.323±0.035 & 0.395±0.018 & \cellcolor{poorbg}0.731±0.04 & 0.534±0.07 & \cellcolor{poorbg}0.856±0.02\\
EA & \underline{\cellcolor{secondbg}}0.297±0.017 & \cellcolor{bestbg}\textbf{0.291±0.027} & 0.464±0.014 & 0.339±0.064 & 0.513±0.023 & \cellcolor{poorbg}0.867±0.07 & 0.715±0.09 & 0.858±0.06\\
$\chi$ & 0.258±0.027 & \cellcolor{bestbg}\textbf{0.235±0.025} & 0.309±0.005 & \underline{\cellcolor{secondbg}}0.256±0.019 & 0.323±0.039 & \cellcolor{poorbg}0.848±0.15 & 0.405±0.02 & 0.784±0.02\\
$\mu_{chem}$ & 0.240±0.011 & \cellcolor{bestbg}\textbf{0.228±0.007} & 0.314±0.017 & \underline{\cellcolor{secondbg}}0.234±0.012 & 0.271±0.018 & \cellcolor{poorbg}0.963±0.14 & 0.419±0.03 & 0.775±0.03\\
$\eta$ & 0.124±0.010 & \underline{\cellcolor{secondbg}}0.113±0.008 & 0.208±0.038 & \cellcolor{bestbg}\textbf{0.101±0.010} & 0.299±0.054 & 0.433±0.04 & 0.331±0.06 & 0.239±0.03\\
$S$ & 56.1±38.9 & 90.0±44.5 & 122.4±41.5 & \underline{\cellcolor{secondbg}}53.3±40.0 & 86.7±23.4 & 438.9±29.8 & 122.5±27.8 & \cellcolor{bestbg}\textbf{51.7±42.2}\\
$\omega$ & 844±611 & 2173±1503 & 1935±846 & \underline{\cellcolor{secondbg}}784.9±696 & 1888±720 & \cellcolor{poorbg}8997±1994 & 2264±747 & \cellcolor{bestbg}\textbf{757.5±739}\\
\midrule
\multicolumn{9}{c}{\emph{Molecular, Geometric, and Charge Properties}}\\
\midrule
$\|\boldsymbol{\mu}\|$ & \underline{\cellcolor{secondbg}}0.638±0.035 & \cellcolor{bestbg}\textbf{0.588±0.039} & 0.737±0.043 & 0.618±0.035 & 0.836±0.13 & 0.857±0.13 & 0.878±0.03 & \cellcolor{poorbg}1.66±0.13\\
$\mu_z$ & 1.43±0.03 & 1.22±0.19 & \cellcolor{bestbg}\textbf{0.888±0.09} & \underline{\cellcolor{secondbg}}1.17±0.05 & 1.32±0.10 & 1.54±1.04 & 2.21±0.12 & 2.14±0.08\\
$r$ & \cellcolor{bestbg}\textbf{0.008±0.001} & \cellcolor{poorbg}0.105±0.016 & 0.070±0.012 & \cellcolor{poorbg}0.141±0.023 & \underline{\cellcolor{secondbg}}0.031±0.006 & 0.292±0.06 & 0.075±0.01 & \cellcolor{poorbg}0.425±0.03\\
$q_\mathrm{maxabs}$ & 0.055±0.005 & \cellcolor{bestbg}\textbf{0.052±0.006} & 0.059±0.003 & \underline{\cellcolor{secondbg}}0.054±0.003 & 0.071±0.010 & 0.067±0.022 & 0.072±0.005 & \cellcolor{poorbg}0.128±0.003\\
$q_\mathrm{absmean}$ & 0.055±0.005 & \cellcolor{bestbg}\textbf{0.052±0.006} & 0.059±0.001 & 0.054±0.003 & 0.078±0.008 & \underline{\cellcolor{secondbg}}0.052±0.007 & 0.072±0.005 & \cellcolor{poorbg}0.129±0.002\\
$q_\mathrm{std}$ & 0.055±0.005 & \cellcolor{bestbg}\textbf{0.052±0.006} & 0.059±0.002 & \underline{\cellcolor{secondbg}}0.054±0.003 & 0.070±0.004 & 0.083±0.013 & 0.072±0.005 & \cellcolor{poorbg}0.127±0.007\\
\bottomrule
\end{tabular}}
\end{table}

\section{Limitations}
While \textit{QuantumCanvas} establishes a rigorous foundation for modeling two-body quantum interactions, several limitations remain. 
First, the dataset is confined to diatomic systems and therefore cannot capture higher-order quantum effects such as collective electron correlation, polarization chains, or cooperative many-body phenomena that arise in extended materials. 
Second, although the ten-channel image representation encodes rich spatial and orbital structure, it remains static: time evolution, excited-state dynamics, and spin-resolved effects are not explicitly represented. 
Third, despite covering 75 elements and 2,850 unique dimers, the dataset does not include relativistic or high-pressure regimes where heavy-element and multi-reference effects become dominant. 
Finally, while pretraining on two-body interactions enhances transfer to larger molecules and crystals, the scalability of this approach to complex condensed phases, for example liquids, alloys, or solid-to-solid transitions, remains an open question. 
We view \textit{QuantumCanvas} as a first step toward a hierarchical and multimodal language of quantum interactions, motivating future extensions to triatomic and many-body systems that incorporate time-dependent, spin-orbit, and external-field effects into unified learning frameworks.

\section{Conclusion}
We introduced \textit{QuantumCanvas}, a large-scale multimodal benchmark that reimagines two-body quantum systems as the fundamental building blocks of matter. 
Across 75 elements and 2,850 unique dimers, the dataset provides 18 scalar quantum descriptors and ten physically grounded image channels derived from orbital populations and charge distributions. 
These representations encode spatial, angular, and electrostatic symmetries that can be directly exploited by modern geometric and vision architectures, bridging data-driven learning and first-principles physics. 
Empirical studies demonstrate that models pretrained on \textit{QuantumCanvas} achieve faster convergence and lower error across larger datasets, with gains up to 25\% prediction accuracy. 
Beyond benchmarking, \textit{QuantumCanvas} establishes a new paradigm for quantum representation learning by representing interatomic physics through interpretable visual and numerical modalities that generalize across molecular, crystalline, and condensed-matter domains. 
As a natural next step, we aim to extend \textit{QuantumCanvas} to triatomic and time-dependent datasets using spin-resolved and relativistic channels, advancing toward universal, physically interpretable foundation models for quantum materials.
{
    \small
    \bibliographystyle{unsrt}
    \bibliography{main}
}

\clearpage
\setcounter{page}{1}
\maketitlesupplementary

\section{Detailed Materials Simulations}
\label{sec:supp-detailed-materials}

Following the finite-temperature Kohn--Sham formulation used in the main paper, each dimer is treated as an isolated two-body system within the self-consistent-charge density-functional tight-binding (SCC--DFTB) approximation at electronic temperature $T_e$~\cite{aradi2007dftb+,gaus2011dftb3,cui2024ptbp}. For completeness, we summarize the finite-temperature occupations, entropy, and the second-order DFTB energy functional that underlies all reported scalar targets.

The single-particle orbitals $\{\psi_i\}$ obey the generalized eigenvalue problem
\begin{equation}
\sum_{\nu} \bigl( H_{\mu\nu} - \varepsilon_i S_{\mu\nu} \bigr) c_{\nu i} = 0,
\end{equation}
where $H_{\mu\nu}$ and $S_{\mu\nu}$ are the Hamiltonian and overlap matrix elements in a localized atomic basis and $c_{\nu i}$ are molecular-orbital coefficients. At finite electronic temperature $T_e$ the orbital occupations are given by Fermi--Dirac statistics
\begin{equation}
f_i = \frac{1}{1 + \exp\!\left( \frac{\varepsilon_i - \mu_F}{T_e} \right)},
\qquad
\sum_i f_i = N_e ,
\end{equation}
with Fermi level $\mu_F$ chosen to conserve the total number of electrons $N_e$.

The corresponding electronic entropy is
\begin{equation}
S[\{f_i\}] = - \sum_i \Bigl[ f_i \ln f_i + (1-f_i)\ln(1-f_i) \Bigr],
\end{equation}
so that the Mermin free energy at fixed nuclear geometry $\mathbf{R} = (\mathbf{R}_A,\mathbf{R}_B)$ reads
\begin{equation}
F[\rho] = E_\mathrm{KS}[\rho] - T_e S[\{f_i\}] ,
\end{equation}
consistent with the expression used in Sec.~3.1 of the main text.

Within SCC--DFTB, the Kohn--Sham energy is approximated by a second-order expansion around a reference density built from neutral atoms. Introducing charge fluctuations $\Delta q_A = q_A - q_A^{(0)}$ relative to the neutral reference, the total energy can be written as
\begin{equation}
E_\mathrm{DFTB}(\mathbf{R},\{\Delta q_A\}) =
E_\mathrm{band}
+ \Delta E_\mathrm{Coul}(\{\Delta q_A\})
+ E_\mathrm{rep}(\mathbf{R}),
\end{equation}
where the band energy is
\begin{equation}
E_\mathrm{band} = \sum_i f_i \varepsilon_i ,
\end{equation}
the short-range repulsive contribution $E_\mathrm{rep}$ is given by a sum of tabulated pair potentials, and the charge-fluctuation term is approximated as
\begin{equation}
\Delta E_\mathrm{Coul}(\{\Delta q_A\})
\simeq
\frac{1}{2}\sum_{A,B} \gamma_{AB}\,\Delta q_A \Delta q_B .
\end{equation}
Here $\gamma_{AB}$ is the chemical hardness matrix derived from the underlying DFT reference and tabulated in the PTBP parameter set.

Mulliken-like gross atomic charges are computed from the density matrix
\begin{equation}
P_{\mu\nu} = \sum_i f_i c_{\mu i} c_{\nu i}
\end{equation}
and the overlap matrix according to
\begin{equation}
q_A = \sum_{\mu \in A} \sum_{\nu} P_{\mu\nu} S_{\mu\nu},
\end{equation}
where the sum over $\mu$ runs over basis functions centered on atom $A$. The SCC correction enters the Hamiltonian matrix elements as
\begin{equation}
H_{\mu\nu}
=
H^{(0)}_{\mu\nu}
+ \frac{1}{2} S_{\mu\nu}
\sum_C \bigl( \gamma_{AC} + \gamma_{BC} \bigr)\,\Delta q_C,
\qquad
\mu \in A,\ \nu \in B ,
\end{equation}
so that $H_{\mu\nu}$, the eigenvalues $\varepsilon_i$, the occupations $\{f_i\}$, and the charges $\{q_A\}$ have to be determined self-consistently.

In practice, each diatomic calculation proceeds by iterating the coupled SCC and SCF equations until both the charge fluctuations and the total-energy change fall below prescribed tolerances,
\begin{equation}
\max_A |\Delta q_A^{(n)} - \Delta q_A^{(n-1)}| < \varepsilon_\mathrm{SCC},
\qquad
|E_\mathrm{tot}^{(n)} - E_\mathrm{tot}^{(n-1)}| < \varepsilon_\mathrm{SCF},
\end{equation}
after which a one-dimensional geometry relaxation along the internuclear separation $r = \|\mathbf{R}_B - \mathbf{R}_A\|$ is performed until
\begin{equation}
\left| \frac{\partial E_\mathrm{tot}}{\partial r} \right| < \varepsilon_\mathrm{geom}.
\end{equation}
All diatomic energies, orbital levels, charges, and dipole moments included in the QuantumCanvas benchmark are evaluated at these self-consistent, geometry-relaxed conditions. The Slater--Koster integrals $H^{(0)}_{\mu\nu}(R_{AB})$, $S_{\mu\nu}(R_{AB})$, the hardness matrix $\gamma_{AB}$, and the repulsive potentials $E_\mathrm{rep}$ are all taken from the periodic-table baseline parameters (PTBP) and evaluated using the \textsc{DFTB+} code~\cite{aradi2007dftb+,gaus2011dftb3,cui2024ptbp}.

\section{Experiment Details}
\label{sec:experimentDetails}

\subsection{Training on QuantumCanvas}
\label{sec:twobody_models}

For the two-body experiments on \textit{QuantumCanvas}, we implement a family of
graph-based, vision-based, and multimodal regressors that all operate on the
same diatomic input representation. Each data point is a two-atom system
$A$–$B$ with equilibrium positions $\mathbf{R}\in\mathbb{R}^{2\times 3}$,
atomic numbers $Z\in\mathbb{N}^2$, a 10-channel $32\times 32$ orbital image
tensor, and a scalar target $y$ (one of the 18 quantum properties). All models
are implemented in PyTorch \cite{imambi2021pytorch} and PyTorch Geometric \cite{fey2019fast} and are trained as scalar
regressors.

\paragraph{Graph-based encoders.}
\textbf{SchNet} follows the reference PyTorch Geometric implementation, with a
lightweight configuration tailored to the two-body setting. We use hidden
channels and filter size of 16, two interaction blocks, eight Gaussian radial
basis functions, and a cutoff radius of $5.0$~\AA. The readout is a simple
atom-wise sum, yielding a single scalar prediction per dimer.

\textbf{DimeNet++} is instantiated with hidden dimension $128$, four interaction
blocks, intermediate embedding sizes of 64 (interaction) and 8 (basis),
out-embedding dimension 256, six radial basis functions, and seven spherical
harmonics. We use a cutoff of $5.0$~\AA{} with an envelope exponent of 5, one
pre-skip and two post-skip layers in each block, and three output MLP layers,
closely following standard DimeNet++ hyperparameters but in a moderately sized
regime suitable for a small-atom system.

\textbf{FAENet} is used in its message-passing configuration with a cutoff of
$5.0$~\AA, eight Gaussian basis functions, 32 hidden channels and filters, and
two interaction layers. Physics-aware embeddings are disabled
(\texttt{phys\_embeds=False}) so that the model focuses on learned
representations rather than engineered descriptors. To accommodate heavy
elements (up to Th, $Z{=}90$), we extend FAENet’s internal embedding tables
(main atomic embedding and optional period/group embeddings) from their default
maximum atomic number to support up to $Z{=}119$ by copying existing weights and
randomly initializing the additional entries with small variance.

\textbf{EGNN-style GCN} is instantiated as a simple isotropic message-passing
baseline. We embed atomic numbers with a 100-way embedding table to a feature
dimension of 128. A $k$-nearest-neighbor graph with $k{=}5$ is constructed from
the 3D coordinates, and three stacked GCNConv layers (all 128 channels) are
applied with ReLU activations. Global graph features are obtained via mean
pooling over nodes, followed by a three-layer MLP (two hidden layers with
ReLU and dropout, and a final linear layer) that maps from 128 to a scalar.

\textbf{GATv2} uses the same $k$-NN graph construction ($k{=}5$) and a 100-way
atomic embedding table to 64 dimensions. We apply three GATv2Conv layers with
4 attention heads, no concatenation (head outputs averaged), and dropout 0.1
in both attention and intermediate activations. Node features are pooled with
global mean pooling and passed through a linear layer to produce the scalar
output. This architecture explicitly learns attention weights over pairwise
interactions conditioned on local geometry.

\paragraph{Vision-based encoders.}
\textbf{ViTRegressor} is a lightweight Vision Transformer operating on
$32\times 32$ images. For models that only consume images, we collapse the
10-channel tensor to 3 channels by selecting the first three channels as a
pseudo-RGB representation. Patches of size $4\times 4$ yield 64 patch tokens,
each embedded via a convolutional patch embedding layer into a 128-dimensional
space. A learnable class token and learnable positional embeddings
for 65 tokens (1 class + 64 patches) are added before feeding the sequence to a
Transformer encoder with 4 encoder blocks, 4 self-attention heads, and a
feed-forward dimension of $4\times 128$. The class token output is passed through
a layer-normalization and a final linear layer to produce the scalar prediction.

\textbf{QuantumShellNet} is a compact CNN specialized for the full
10-channel orbital image tensor. Five convolutional layers with stride~2
and kernel size~3 progressively downsample the $32\times 32$ input to a
$1\times 1$ spatial representation:
$10\rightarrow 80\rightarrow 160\rightarrow 96\rightarrow 48\rightarrow 24$
channels. Each convolution is followed by ReLU and dropout (0.3).
The resulting 24-dimensional latent vector is fed into a stack of fully
connected layers interleaved with coarse atomic descriptors. In particular,
we approximate per-dimer scalar features: total proton number (sum of $Z$),
approximate mass number (using $A\approx 2Z$ for each atom), and total neutron
number ($A{-}Z$). These three scalars are concatenated at multiple stages of the
MLP (after the first and second hidden layers) to explicitly inject basic
nuclear information. The final MLP has hidden sizes 200, 100, and 50, with
ReLU and dropout, culminating in a scalar regression head. All convolutions and
linear layers operate in single-precision (float32).

\paragraph{MultiModal}
Combines SchNet-derived geometric features with
a ResNet-18 backbone acting on the orbital images. On the graph side, we use
the same SchNet configuration as above (16 hidden channels, two interactions,
cutoff $5.0$~\AA). Rather than using only the final scalar output, we extract
atom-wise hidden features after the last interaction block, and apply global
mean pooling over atoms to obtain a $16$-dimensional graph-level descriptor.
On the vision side, we use torchvision’s ResNet-18 pretrained on ImageNet,
remove the final classification fully-connected layer, and treat the penultimate
$512$-dimensional global average pooling vector as the image descriptor. The
orbital tensor is again reduced to 3 channels by using the first three channels
for compatibility with the pretrained weights.

The two modality-specific embeddings are concatenated into a
$(16{+}512)=528$-dimensional joint representation and passed through a
three-layer MLP with hidden widths $256 \rightarrow 128 \rightarrow 64$, ReLU
activations, and dropout $0.1$ after each hidden layer. A final linear layer
maps the fused representation to a scalar prediction. This architecture
corresponds to a late-fusion baseline in which the network must learn how to
weight geometric and image-based cues without explicit cross-attention.

\paragraph{Dataset interface and label normalization.}
All models consume data through a unified \texttt{QuantumCanvas} wrapper built
on \texttt{torch\_geometric.data.Data}. Each sample stores atomic numbers $z$,
Cartesian coordinates $\mathbf{r}$, the 10-channel image tensor (if available),
and a scalar label $y$. For training stability, target values are min–max
normalized to the range $[-1,1]$ using statistics computed on the training
split only; the same statistics are reused for validation and test splits, and
all metrics are reported after denormalizing predictions back to the original
physical units.

All models are trained under a unified and fully reproducible benchmarking
pipeline implemented in PyTorch and PyTorch Geometric. Each experiment is
specified by a target property $y$ and a random seed $s$, and all runs follow
the same data handling, optimization scheme, evaluation protocol, and logging
infrastructure.

\paragraph{Dataset construction and splits.}
For every experiment, we first instantiate the \texttt{QuantumCanvas} to read
the full QuantumCanvas dataset and sample count $N$. Indices are randomly
shuffled using a seed-controlled \texttt{torch.Generator}, and the dataset is
partitioned into an 80/10/10 split:
\[
\mathcal{D}_{\text{train}},\quad
\mathcal{D}_{\text{val}},\quad
\mathcal{D}_{\text{test}}.
\]
Normalization statistics (min/max for the supervised target) are computed
\emph{only} from $\mathcal{D}_{\text{train}}$, and reused for validation and
test. Each \texttt{Data} object contains atomic numbers $z\in\mathbb{N}^2$,
positions $\mathbf{r}\in\mathbb{R}^{2\times 3}$, the 10-channel orbital image
tensor $\mathbf{I}$, and the normalized label $\hat{y}$.

\paragraph{Input handling across model classes.}
The training loop dispatches data to each architecture according to its
expected interface:
\begin{itemize}
    \item \textbf{SchNet, DimeNet++, GATv2, EGNN} consume $(z, \mathbf{r}, \text{batch})$.
    \item \textbf{FAENet} receives a full PyG batch object and may internally
    return either a tensor or a dictionary; the regression head is extracted
    accordingly.
    \item \textbf{ViTRegressor} and \textbf{MultiModal} use the first
    three channels of $\mathbf{I}$ as a pseudo-RGB image (for compatibility with
    ImageNet-pretrained ResNet-18 in the multimodal setting).
    \item \textbf{QuantumShellNet} processes the full 10-channel
    $32\times32$ tensor and also incorporates $(z, \mathbf{r}, \text{batch})$
    information in its fully connected layers.
\end{itemize}

\paragraph{Training loop.}
All models are optimized using the loss function specified in the configuration
(\texttt{MAE} or \texttt{MSE}). For a mini-batch $\{(\mathbf{x}_i,\hat{y}_i)\}$,
the model produces predictions $\hat{y}_i^{\,\text{pred}}$, the loss is
backpropagated, and parameters are updated using one of the configured
optimizers (Adam, AdamW, or SGD):
\[
\mathcal{L} = 
\begin{cases}
\frac{1}{B}\sum_i |\hat{y}_i - \hat{y}_i^{\,\text{pred}}| & \text{MAE},\\[4pt]
\frac{1}{B}\sum_i (\hat{y}_i - \hat{y}_i^{\,\text{pred}})^2 & \text{MSE}.
\end{cases}
\]
Learning rate schedules follow the configuration:
ReduceLROnPlateau (validation-loss monitored), cosine annealing, step-decay, or
no scheduling. Training continues for up to 50 epochs with early stopping
(patience = 10).

\paragraph{Evaluation and denormalization.}
During validation and testing, predictions are first computed in the normalized
space and then mapped back to physical units via the train-set statistics:
\[
y = y_{\min} + \frac{\hat{y}+1}{2} (y_{\max}-y_{\min}).
\]
We report MAE and RMSE in the original physical units (eV, Debye, Å, etc.).
The checkpoint with the lowest validation loss is used for final test
evaluation. Random seeds are applied across PyTorch, NumPy, and dataset index permutation to
ensure exact reproducibility.

\paragraph{Benchmark coverage.}
The full benchmark spans \emph{all} models listed in
Sec.~\ref{sec:twobody_models}, and 22 supervised targets drawn from electronic,
energetic, geometric, dipole, and charge-related properties. For each target we
run experiments across three independent seeds, totaling $22\times 3$ supervised
tasks and training each of the 8 architectures under identical conditions.

\subsection{Fine-tuning on downstream benchmarks}
\label{sec:downstream_finetune}

To assess how two-body representations learned on \textit{QuantumCanvas} transfer
to larger and more complex systems, we fine-tune a subset of our architectures
on three standard downstream benchmarks: \textit{QM9}, \textit{MD17}, and
\textit{CrysMTM}. Throughout, we focus on two strong graph-based encoders,
\textbf{SchNet} and \textbf{GotenNet}, which are widely used in molecular and
materials machine learning and provide clean, scalable baselines for studying
cross-domain transfer. In all cases we reuse exactly the same architectural
configurations as in the two-body setting so that \textit{QuantumCanvas}
checkpoints can be loaded without any structural changes.

Across all benchmarks we adopt a unified training protocol wherever possible:
mini-batches of size $32$, Adam optimization with weight decay $0.0$, a
ReduceLROnPlateau scheduler (factor $0.8$, patience $10$, minimum learning rate
$10^{-6}$), and early stopping based on validation MAE. For each dataset we
run three seeds ($42$, $123$, $456$), and for every (target, seed) pair we
train two variants per architecture: a \emph{from-scratch} baseline initialized
with random weights (learning rate $10^{-4}$), and a \emph{fine-tuned} model
initialized from the corresponding two-body \textit{QuantumCanvas} checkpoint
and updated with a smaller learning rate $10^{-5}$ to mitigate catastrophic
forgetting.

\paragraph{QM9: frontier orbital properties.}
On \textit{QM9} we target three central electronic quantities: the HOMO energy,
the LUMO energy, and the HOMO–LUMO gap. We use the standard PyG split with
$110{,}000$ molecules for training, $10{,}000$ for validation, and $10{,}000$
for testing. For each target and seed we train both \textbf{SchNet} and
\textbf{GotenNet} under the two regimes described above. Fine-tuned runs are
initialized from two-body checkpoints trained on the matched
\textit{QuantumCanvas} targets (\texttt{e\_homo\_ev}, \texttt{e\_lumo\_ev},
\texttt{e\_g\_ev}). The QM9 models use the same SchNet and GotenNet
hyperparameters as in Sec.~\ref{sec:twobody_models}
(96 hidden channels and six interaction blocks for SchNet; 64-dimensional atom
features and three interaction blocks for GotenNet), ensuring that any
performance gains can be attributed to two-body pretraining rather than changes
in model capacity.

Training uses an L1 loss on the relevant component of the QM9 label vector.
All optimization settings follow the shared protocol above, with the only
difference between scratch and fine-tuning runs being the learning rate
($10^{-4}$ vs.\ $10^{-5}$). For each (target, seed, regime) we select the
checkpoint with the lowest validation MAE and report MAE and RMSE on the test
set in the original QM9 units.

\paragraph{MD17: molecular dynamics energies.}
On \textit{MD17} we probe transfer to atomistic molecular dynamics by training
on three canonical molecules: benzene, aspirin, and ethanol. For each system we
construct a compact yet representative split of $950$ geometries for training,
$50$ for validation, and a held-out test subset of $1{,}000$ conformations.
Both \textbf{SchNet} and \textbf{GotenNet} are trained to predict the scalar
potential energy of each frame, again using the two-body architectures without
modification (SchNet with 96 hidden channels, six interaction blocks, 50 radial
basis functions and cutoff $5.0$~\AA; GotenNet with 64-dimensional atom
features, three interactions, 10 radial basis functions, and $5.0$~\AA{}
cutoff). Fine-tuned runs are initialized from the two-body
\texttt{total\_energy\_ev} checkpoint.

For each molecule, targets are normalized using training-set statistics: we
compute the mean and standard deviation of the energy over the $950$ training
frames and optimize an L1 loss in this normalized space. During evaluation,
predictions are mapped back to the original MD17 energy units and we report MAE
and RMSE on the denormalized values. Optimization hyperparameters follow the
QM9 configuration (batch size $32$, up to $50$ epochs, Adam, scheduling, and
early stopping as above), with the same distinction between scratch and
fine-tuning learning rates. The best checkpoint per (molecule, seed, regime) is
selected by validation MAE and used for final test reporting.

\paragraph{CrysMTM: crystalline frontier-orbital energies.}
On the \textit{CrysMTM} TiO$_2$ benchmark we examine transfer from diatomic
pretraining to periodic oxide materials. The dataset consists of TiO$_2$
crystals sampled across 21 temperatures (0–1000~K in 50~K increments) and
multiple rotational configurations. We use all available configurations at
these temperatures and focus on two frontier-orbital quantities, \textbf{HOMO}
and \textbf{LUMO}, which are most directly aligned with the electronic targets
in \textit{QuantumCanvas} and thus form a clean testbed for cross-domain
transfer. As in QM9 and MD17, we use \textbf{SchNet} and \textbf{GotenNet} with
architectures fixed to their two-body configurations, enabling weight reuse
without any structural modification.

For each property we randomly partition the dataset into an 80/20 split using a
seed-controlled permutation at the crystal level, train on the 80\% subset, and
hold out the remaining 20\% for evaluation. Targets are standardized per
property using the mean and standard deviation computed on the training subset,
and models are optimized with an MSE loss in this normalized space. All
reported metrics (MAE and RMSE) are computed after denormalizing predictions
back to the original physical units (eV). Optimization settings mirror the
other benchmarks, with the exception that we allow up to $100$ epochs and use a
single base learning rate of $10^{-4}$ (and $10^{-5}$ for fine-tuning), and
early stopping patience is reduced to $20$ validation steps.

For each target (HOMO, LUMO) and each seed we train from-scratch and
fine-tuned variants of both architectures, where fine-tuned runs are
initialized from the corresponding two-body \textit{QuantumCanvas} checkpoints
(\texttt{e\_homo\_ev} for HOMO and \texttt{e\_lumo\_ev} for LUMO). This yields
$2 \times 3 \times 2 = 12$ CrysMTM experiments per model, carried out under the
same data handling, normalization, and optimization protocol as the molecular
benchmarks, enabling a controlled comparison of two-body pretraining across
molecular, dynamical, and crystalline regimes.

\end{document}